\documentclass[runningheads]{llncs}
\usepackage{graphicx}
\usepackage{comment}
\usepackage{amsmath,amssymb} % define this before the line numbering.
\usepackage{color}
\usepackage{eso-pic}
\usepackage{multirow}
\usepackage{subfigure}

\usepackage[width=122mm,left=12mm,paperwidth=146mm,height=193mm,top=12mm,paperheight=217mm]{geometry}

\begin{document}
% \linenumbers
\pagestyle{headings}
\mainmatter
\def\ECCVSubNumber{100}  % Insert your submission number here

\title{MSDU-net: A Multi-Scale Dilated U-net for Blur Detection} % Replace with your title

%\begin{comment}
\titlerunning{ECCV-20 submission ID \ECCVSubNumber}
\authorrunning{ECCV-20 submission ID \ECCVSubNumber}
\author{Anonymous ECCV submission}
\institute{Paper ID \ECCVSubNumber}
%\end{comment}
%******************

%\begin{comment}
\titlerunning{Abbreviated paper title}
\author{Fan Yang\inst{1} \and
Xiao Xiao\inst{2}}
\authorrunning{F. Author et al.}
\institute{Xidian University, Province Shanxi, Chnia \email{fyang\_jfsl@stu.xidian.edu.cn} \and
Xidian University, Province Shanxi, Chnia \email{xiaoxiao@xidian.edu.cn}}
%\end{comment}
%******************
\maketitle

\begin{abstract}
Blur detection is the separation of blurred and clear regions of an image, which is an important and challenging task in computer vision. In this work, we regard blur detection as an image segmentation problem. Inspired by the success of the U-net architecture for image segmentation, we design a Multi-Scale Dilated convolutional neural network based on U-net, which we call MSDU-net. The MSDU-net uses a group of multi-scale feature extractors with dilated convolutions to extract texture information at different scales. The U-shape architecture of the MSDU-net fuses the different-scale texture features and generates a semantic feature which allows us to achieve better results on the blur detection task. We show that using the MSDU-net we are able to outperform other state of the art blur detection methods on two publicly available benchmarks.
\end{abstract}

\section{Introduction}

	Image blurring comes in two main flavors: defocus blur, which is caused by defocusing and motion blur, which is caused by camera or object motion. Blur detection aims to detect the blurred regions in an image regardless of the cause and separate them from sharp regions. This task plays a significant part in many potential applications, such as salient object detection \cite{Chang2016Salient,sun2017focus}, defocus magnification \cite{Chang2013Defocus,bae2007defocus}, image quality assessment \cite{tang2017effective,wang2008blind}, image deblurring \cite{shi2014discriminative,masia2011coded}, image refocusing \cite{zhang2009single,zhang2011single} blur reconstruction \cite{wang2016allfocus,munkberg2016layered}, \textit{etc.}
	
	\begin{figure}[htbp]
		\centering

        \subfigure[]{	
			\begin{minipage}{0.25\linewidth}	
				\includegraphics[width=1.0\linewidth]{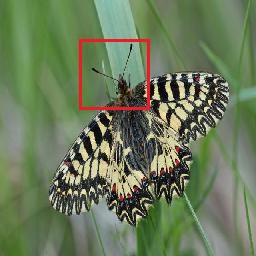}
			\end{minipage}%	
		}%
		\subfigure[]{	
			\begin{minipage}{0.25\linewidth}	
				\includegraphics[width=1.0\linewidth]{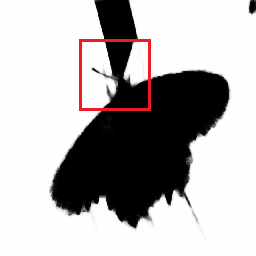}
			\end{minipage}%	
		}%
		\subfigure[]{
			\begin{minipage}{0.25\linewidth}
				\includegraphics[width=1.0\linewidth]{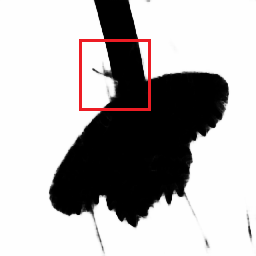}
			\end{minipage}	
		}%
		\subfigure[]{
			\begin{minipage}{0.25\linewidth}	
				\includegraphics[width=1.0\linewidth]{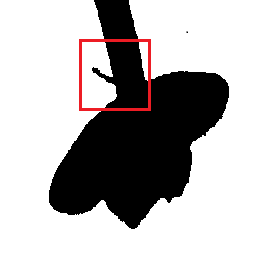}
			\end{minipage}	
		}%
		\centering
		
		\caption{(a) is the source image; (b) is the result of U-net; (c) is the result of our method; (d) is the ground truth. We can see that the U-net is good at blur detection. And our method makes progress on the baseline.}	
		\label{fig1}
	\end{figure}
	
	In the past decades, a series of blur detection methods based on hand-crafted features have been proposed. These methods exploit various hand-crafted features related to the image gradient \cite{yi2016lbp,golestaneh2017spatially,karaali2017edge,xu2017estimating}, and frequency \cite{shi2014discriminative,shi2015just,tang2016spectral,Chang2013Defocus}. These method tend to measure the amount of the feature information contained in different image regions to detect blurriness since the blurred regions usually contain fewer details than the sharp ones. However, these hand-crafted features are usually not good at differentiating the sharp regions from the complex background, and cannot understand semantics to extract sharp regions from a similar background.
	
	\begin{figure}[thbp]		
		\centering
		\includegraphics[width=1.0\linewidth]{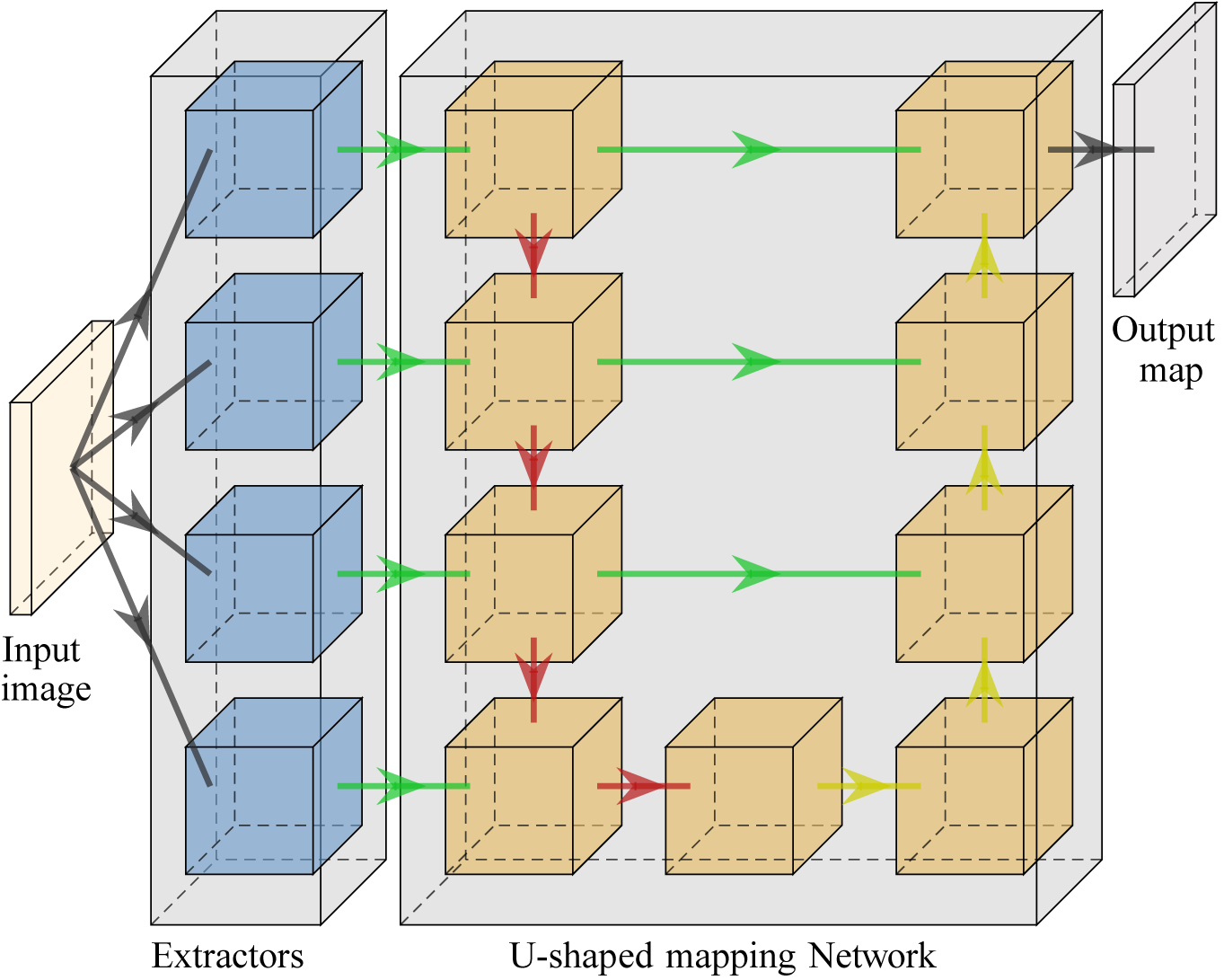}
		\caption{A diagram of our whole network, gray arrows are normal data flows; green arrows are copy and connection; red arrows are contracting with pooling layers; yellow arrows are expanding with up-sampling layer; blue blocks are extractors with dilated convolution; orange blocks are common convolution blocks.}
		\label{fig2}
	\end{figure}

	Recently, deep convolutional neural networks(DCNNs) have made amazing progress in machine learning and artificial intelligence. It has made vital contributions to various computer vision tasks, such as image classification \cite{he2016deep,huang2017densely}, object detection \cite{ren2015faster,redmon2018yolov3} and tracking \cite{qi2016hedged,li2018deep}, image segmentation \cite{zhao2017pyramid,ronneberger2015u}, image denoising \cite{jin2017deep,zhang2017beyond} and super resolution \cite{kim2016accurate,dong2015image} and \textit{etc.}. Several DCNNs-based methods have been proposed to address blur detection. They both used a lot of various extractors to capture the essential feature information to detect blur information.
	
    From a different perspective, we regard blur detection as an image segmentation problem, which can simplify the blur detection problem to blur features extracting and blur feature mapping. Learning from the classical fully convolutional networks (FCNs) architectures \cite{long2015fully} applied in image segmentation, we propose a feature mapping network based on U-net \cite{ronneberger2015u} with a group extractors using dilated convolution, which can extract the features on multi-scales.

    Instead of using a number of different feature extractors in Cross-Ensemble Network \cite{zhao2019enhancing} and DeFusionNet \cite{tang2019defusionnet}, we design a group of extractors to capture texture information for blur detection. Based on the study of dilated convolution kernels \cite{chen2019drop,wang2018understanding}, we use the dilated convolution with various dilation rates and strides to capture the texture information from different scales. Specifically, we use the low dilation convolution and small stride convolution to capture the texture information from small scale, and use the high dilation rate convolution and big stride convolution to capture the texture information from large scale. We find out that texture information can describe the degree of blur, and semantic information can help judge whether the region is blurred. Thus, we propose a new U-shaped network based on \cite{ronneberger2015u} to integrate different scale texture features and semantic information for blur detection. Our model consists of different-scale features extractors and a U-shaped network, which shows in Figure \ref{fig2}.

    \begin{comment}
	From the previous achievement, we find out that texture information can describe the degree of blur, and semantic information can help judge whether the region is blurred. Instead of increasing the type and number of feature extractors, we try to design a group of extractors to capture texture for blur detection. Based on the study of dilated convolution kernels \cite{chen2019drop,wang2018understanding}, we try to use the convolution with various dilation rates and strides to capture the texture information from different scales. Specifically, we combine the low dilation convolution and small stride convolution to capture the texture information from small scale, and combine the high dilation rate convolution and big stride convolution to capture the texture information from large scale.
	
	From a different perspective, we simplify the blur detection problem to blur features extracting and blur feature mapping, through which blur detection come to be image segmentation. Learning from the classical fully convolutional networks (FCNs) architectures \cite{long2015fully} applied in image segmentation, we design a feature mapping network based on U-net \cite{ronneberger2015u}, which can integrate different levels of features. Thus, our model consists of different-scale features extractors and a U-shaped network, which shows in Figure \ref{fig2}.	
	\end{comment}
	To sum up, our main contributions follow:
	\begin{itemize}
	    \item We design a new model based on U-net \cite{ronneberger2015u} from image segmentation. The new model can fuse the multi-scale texture information and the semantic information to get more accurate results of blur detection.
	
	    \item We propose a group of extractors with dilated convolution to capture multi-scale texture information on purpose rather than straightly stack many extractors.
	
		\item Our method addresses the blur detection ignoring the specific cause of the blur, which can detect both defocus blur and motion blur.
		
		\item Compared with the state-of-the-art blur detection methods, the proposed model can get the best F$_{\sqrt{0.3}}$-measure scores all over $95\%$ on the three datasets. Moreover, we have done a series of experiments to analyze and verify the effectiveness of the multi-scale extractors and skip connections in our model.
	
	\end{itemize}

\begin{comment}
	In summsary, in this work provides the following contribution:
	\begin{itemize}
		\item To our knowledge, we are the first work to propose using a U-net for blur detection, framing it as an image segmentation problem.
		\item We design a new group of feature extractors with dilated convolution which can efficiently extract different-scale texture information directly, and integrate this with the U-net in a novel manner. This improvement makes the model more accurate and compact.
		\item Our method addresses the blur detection of how to distinguish between the blur regions and clear regions, ignoring the specific cause of the blur. Therefore, our method can detect both defocus blur and motion blur.
		\item Compared with the state-of-the-art blur detection methods, the model proposed by us can get the bestF$_{\sqrt{0.3}}$-measure scores all over $95\%$ on the  three datasets.
	\end{itemize}
\end{comment}
	
\section{Related Work}
	We refer to the previous achievement on the blur detection and the semantic segmentation. We can find the trend of using deep learning networks for the blur detection from the previous achievement. Considering the blur detection as an image segmentation problem, we can learn how to get a more accurate segmentation result from semantic segmentation for the blur detection.
	\subsection{Previous Achievement}
	Previous methods of blur detection can be divided into two categories: methods based on traditional hand-crafted feature and methods based on deep learning neural networks.
	
	In the first category, various hand-crafted features exploit gradient and frequency that can describe the information of regions. For example, Su \textit{et al.} \cite{su2011blurred} combine the gradient distribution pattern of the alpha channel with a blur metric based on singular value distributions to detect the blurred region. Shi \textit{et al.} \cite{shi2014discriminative} make use of a series of the gradient, Fourier domain, and data-driven local filters features to enhance discriminative power for differentiating blurred and un-blurred image regions. Considering that feature extractors based on local information cannot distinguish the just noticeable blur reliably from unblurred structures, Shi \textit{et al.} \cite{shi2015just} improved feature extractors via sparse representation and image decomposition. Yi \textit{et al.} \cite{yi2016lbp} design a sharpness metric based on local binary patterns and the focus and defocus image regions are separated by using the metric. Tang \textit{et al.} \cite{tang2016spectral} design a log averaged spectrum residual metric to obtain a coarse blur map, and propose iterative updating mechanism to refine the blur map from coarse to a fine based on the intrinsic relevance of similar neighbour image regions. Using discrete cosine transform, Golestaneh \textit{et al.} \cite{golestaneh2017spatially} computed blur detection maps based on a high-frequency multi-scale fusion and sort transform of gradient magnitudes.
	
	Due to their outstanding performance in high-level feature extraction and parameters learning, deep convolutional neural networks have reached new state-of-the-art on blur detection. Firstly, Park \textit{et al.} \cite{park2017unified} combine deep patch-level and hand-crafted features together to estimate the degree of defocus. Huang \textit{et al.} \cite{huang2018multiscale} design a patch-level CNN to learn blur features, and apply this net work at three coarse-to-fine scales and optimally fuse multi-scale blur likelihood maps to generate better blur detection. Patch-level DCNN methods are time-consuming, which is needed to run thousands of times to process a raw image. Zhao \textit{et al.} propose a multi-stream bottom-top-bottom fully convolutional network \cite{zhao2018defocus} which integrates low-level cues and high-level semantic information for defocus blur detection and leverages a multi-stream strategy to handle the defocus degree's sensitivity to image scales. In \cite{ma2018deep}, Ma \textit{et al.} exploit the high-level information to separate the blur regions through an end-to-end fully convolution network. In order to increase the efficiency of the network, Tang \textit{et al.} propose a new blur detection deep neural network via recurrently fusing and refining multi-scale features \cite{tang2019defusionnet} . Zhao \textit{et al.} design the Cross-Ensemble Network \cite{zhao2019enhancing} with two groups of defocus blur detectors, which is alternately optimized with cross-negative and self-negative correlation losses to enhance the diversity of features.
	
	With the application of DCNNs in computer vision, more solutions have been proposed for blur detection. Making the network deeper or wider to catch more useful features has been proven applicable, but this way is so dull that it makes unnecessary consumption. In our work, we make an attempt to design a delicate neural network to solve blur detection more efficiently.
	
	\subsection{Image Segmentation}
	
	As we know, fully convolutional networks (FCNs) \cite{long2015fully} which train end-to-end, pixel-to-pixel on semantic segmentation exceed the previous best results without further machinery. Various improved versions of FCNs have been applied to region proposals \cite{ren2015faster}, contour detection \cite{song2019building}, depth regression \cite{fu2018deep}, optical flow \cite{mahendran2018self} and weakly-supervised semantic segmentation \cite{xian2019semantic} which further advance the state-of-the-art in image processing. Some classical architectures have good performances in image segmentation. Representative of them are DeepLab models \cite{chen2017deeplab,chen2017rethinking,chen2018encoder} and U-net \cite{ronneberger2015u}.
	
	 In DeepLab models \cite{chen2017deeplab,chen2017rethinking,chen2018encoder}, dilated convolution and dense conditional random field inference (CRF) are used to improve output resolution. ParseNet \cite{liu2015parsenet} normalizes features for fusion and captures context with global pooling. The "deconvolutional network" has been proposed by \cite{zeiler2010deconvolutional} to restore resolution by proposals stacks of learned deconvolution and unpooling. In \cite{fu2019stacked}, the stacked deconvolutional networks are applied in semantic segmentation, which can get an outstanding result without CRF. Almost all kinds of effective tricks in convolution networks have been applied in FCNs to improve the results of the network.
	
	 U-shape network has been first proposed in \cite{ronneberger2015u} to address biomedical image segmentation, which only has a few training samples. To make best use of the limited samples, U-Net \cite{ronneberger2015u} combines skip layers and learned deconvolution to fuse the different level features of one image for the more precise result. Because of its outstanding performance for the biomedical datasets that has simple semantic information and a few fixed feature, there are many further studies based on it,such as VNet \cite{milletari2016v} that is the U-shaped network using three-dimensional convolutions, UNet++ \cite{zhou2018unet++} that is U-shaped network with more dense skip connections, Attention U-Net \cite{oktay2018attention} that combines U-shaped networks with attention mechanism, ResUNet-a \cite{diakogiannis2019resunet} that implements the U-shaped network with residual convolution blocks, TernausNet \cite{iglovikov2018ternausnet} use the pre-trained encoder to improve the U-shape network, MDU-Net \cite{zhang2018mdu} that densely connects the multi-scale of U-shaped network, and LinkNet \cite{chaurasia2017linknet} that attempts to modify the original U-shaped network for efficiency.

	The achievements of U-shape network provide a lot of valuable references for us to solve blur detection. Considering that blur detection also has a few fixed feature, we design our network based on U-Net\cite{ronneberger2015u}.
	
	\begin{figure*}[thbp]
		\begin{center}
			\centering
			\includegraphics[width=1.0\linewidth]{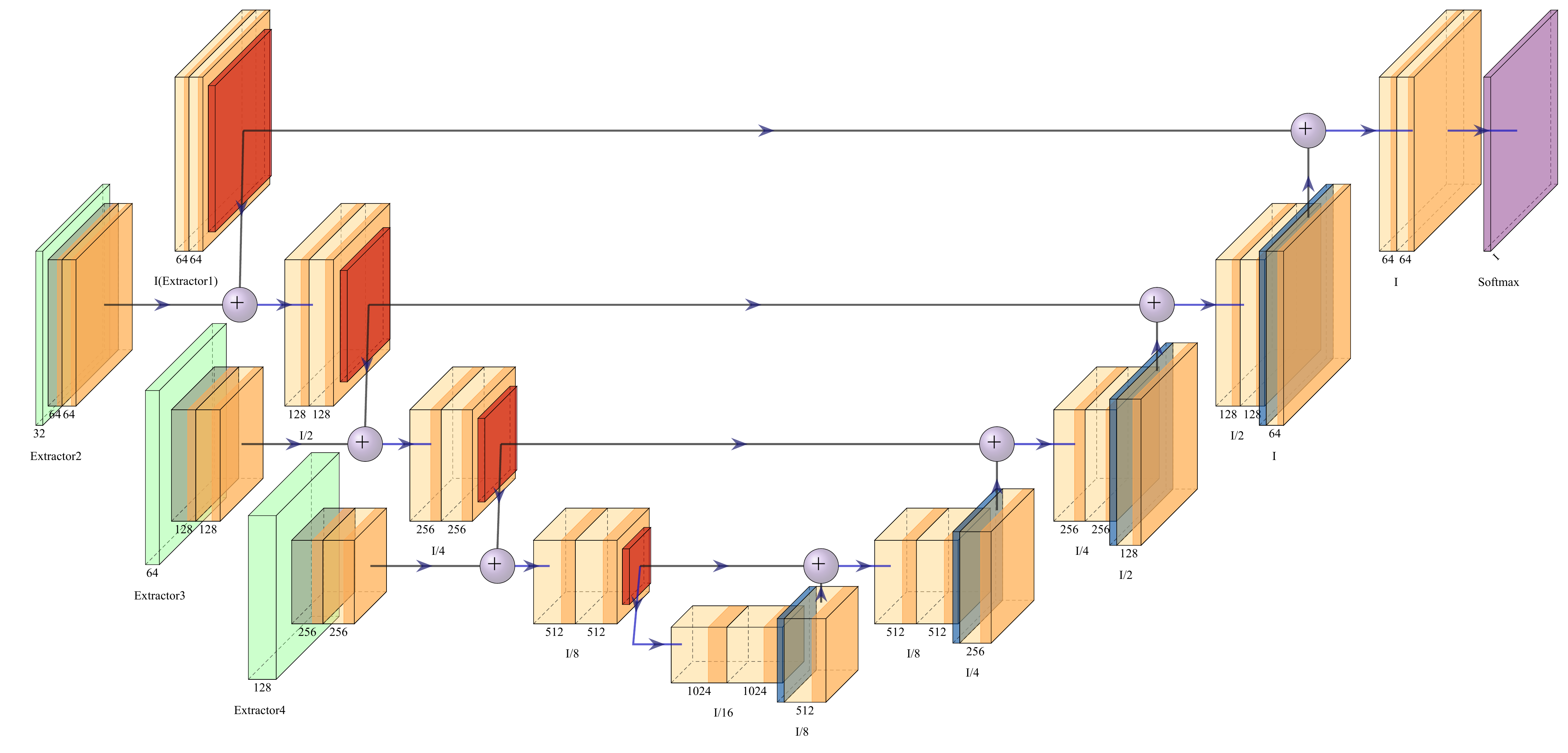}
		\end{center}
		\caption{A detailed diagram of our model. Green blocks are dilated convolutions; red blocks are max pooling layers; orange blocks are convolutions with a Relu activation function; gray blocks are downsampling convolutions with Relu activation functions; blue blocks are upsampling convolutions; purple blocks are softmax layers; "$+$" means that concatenation of the feature matrices with the same height and width in the channel dimension.}
		\label{fig3}
	\end{figure*}
	
\section{Proposed MSD-Unet}
	Our model consists of two parts: a group of extractors and the U-shaped network. First, we use the group of extractors to capture the multi-scale texture information from the images. Then, we put the extracted feature maps in each contracting steps of U-shaped networks to integrate them together. Finally, we use a soft-max layer to map the feature matrix to the segmentation result. The whole model was shown in detail in Figure \ref{fig3}.
	
		\begin{figure}[thbp]	
		\subfigure[Normal $3 \times 3$ convolutional kernel]{
			\begin{minipage}{0.5\linewidth}	
				\centering
				\includegraphics[width=1.0\linewidth]{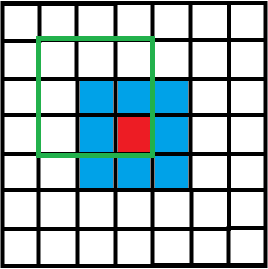}
			\end{minipage}%
		}%
		\subfigure[Dilated $3 \times 3$ convolutional kernel]{
			\begin{minipage}{0.5\linewidth}	
				\centering
				\includegraphics[width=1.0\linewidth]{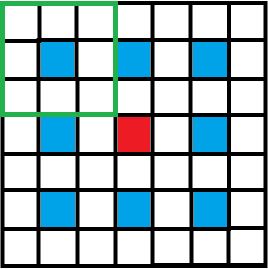}
			\end{minipage}%
		}%
		\caption{The dilated convolutional kernel has bigger receptive field than normal convolutional kernel, as the receptive fields of single pixels are the same.}
		\label{fig4}
	\end{figure}
	
	\subsection{Basic Components}
	To improve the efficiency of the extractors, we use dilated convolution to replace standard convolution, which can enlarge the receptive field without increasing parameters. And why we chose the U-shape network to fuse the different-scale texture information is that the skip connections in U-shape network can concatenate the shallow feature matrices and the deep feature matrices, which help us make best use of texture information and semantic information.

	\textbf{Dilated convolution}
	Dilated convolution, also called atrous convolution, was originally developed in algorithms for wavelet decomposition \cite{holschneider1990real}. The main idea of the dilated convolution is to insert a hole between pixels in the convolutional kernel to increase its receptive field. The dilated convolution can effectively improve the extraction ability of convolution kernels for more features with the fixed number of parameters. In Figure \ref{fig4}, the $3 \times 3$ convolution kernel with dilation of $2$ has a $5 \times 5$ receptive field as big as a $5 \times 5$ convolution kernel does, and has the same number of parameters as the normal $3 \times 3$ convolution kernel. Dilated convolutional kernels have different dilation rates. If we set the center of the convolution kernel as the origin of the coordinates, for a 2-D convolution kernel with size $k \times k$, the result of $r$ dilation can be as following:
	\begin{equation}
	\alpha=r-1
	\label{equa1}
	\end{equation}
	\begin{equation}
	S_{d}=S_{o}+(S_{o}-1)\cdot\alpha
	\label{equa2}
	\end{equation}
	where $S_{d}$ is the size of the dilated convolution kernels, $S_{o}$ is the size of the origin convolution kernel, $\alpha$ is the dilation factor.
	
	\begin{equation}
	K_{d}(x, y) = \begin{cases}
	K_{o}(i, j) & \mathrm{if, } \ x = i \cdot \alpha  , y = j \cdot \alpha \\
	0 & \mathrm{else}
	\end{cases} \\
	\label{equa3}
	\end{equation}
	where $K_{d}(x, y)$ is a single parameter in the dilated convolution kernel, $K_{o}(x, y)$ is a single parameter in the origin convolution kernel. In Figure \ref{fig4}, we can see a $3\time3$ convolutional kernel change to dilated convolutional kernel with $2$ dilation. With the deeplearning method we can use more deeper network to catch the more abstract features. However, whether the region is blured is due to direct features. Thus, We need to increase the receptive field by expanding the size of the convolution kernel without making the deeper. In our method, we exploit dilated convolutions to design a group of extractors which can extract texture information but does not need more additional parameters.
	
	\textbf{Skip connections}
	 Skip connections combine the straight shallow features and abstract deep features, which can make the network notice shallow texture information and deep semantic information to have a more precise result. As we know, the more convolution layers are stacked, the more high-level abstract information can be extracted. Traditional encoder-decoder architectures can extract high-level semantic information, and performs well in panoramic segmentation that contains abundant high-level information. However, if we have to make images segment with the data only containing poor high-level information, such as cell splitting, MIR image segmentation, satellite image segmentation, etc., we should efficiently exploit the low-level information. The skip connections retain the low-level features in the shallow layers and combine them with the high-level features after deep layers, which can make the best use of both high-level and low-level information. For our task, the low-level information of gradient and frequency can describe the absolute degree of blur, and the high-level information of global semantics can help to judge whether the regions are blurred. As a result, the skip connections can make our model robust to adapt to various backgrounds.

	\subsection{Model Details}
	 Our model is based on U-net \cite{ronneberger2015u}, so they have a similar structure. The U-shaped architecture can be seen as having two paths: the contracting path and the expansive path. However, to combine with the multi-scale texture extractors, we have modified the contracting path of the U-net to receive different-scale texture feature matrices in every stage. In this section, we describe the detail of our model from the extractors and U-shape network.
	
	\textbf{Extractor}
	We design the extractors aiming at capturing the multi-scale texture feature. Firstly, the source image is fed in the dilated convolution layers that dilation rates are various of ${1,2,2,2}$ and kernel size is $3 \times 3$. Secondly, we contact the output of the dilated convolution layers with regular convolution layers with strides of ${1,2,4,8}$ with ReLU activation function, which can shrink the size of feature maps to fit the size of the feature of each contracting path in U-shaped architectures. After that, we add another regular convolution layers at the end to smooth the different-scale features. Why we make all the extractors independently is that the independent extractors are contacted contracting path in different levels of the U-shaped architecture can make the model more robust with different scales.
	
	\textbf{U-shaped architecture}
	  The contracting path which receives the outputs of texture extractors and integrates them through concatenation, convolution and pooling makes the feature matrices shrink in length and width dimension and grow in channel dimension. The expansive path uses transposed convolutions to restore the resolution of feature matrices and concatenate them with the feature matrix that has the same size in the contracting path through skip connections. U-shape architecture use skip layers concatenate the feature channels of the two paths in the upsampling part, which allow the network to propagate semantic information to higher resolution layers that contain local texture information.
	
	As a consequence, the expansive path is more or less symmetric to the contracting path, and yields a u-shaped architecture. The contracting path follows the typical architecture of a U-net \cite{ronneberger2015u}, which consists of the repeated application o f two $3\time3$ convolutions, each followed by a rectified linear unit and a $2\time2$ max pooling operation with stride $2$ for downsampling. The input feature maps of every step in the contracting path is combined with the output of the last step and the corresponding extractor. The expansive path is the same as U-net, which consist of a $2\time2$ transposed convolution that halves the number of feature channels, a concatenation with the correspondingly cropped feature map from the contracting path, and two 3x3 convolutions, each followed by a ReLU.

	\section{Experiments}

	\subsection{Dataset and Implementation}
	
	We do our experiments on two publicly available benchmark datasets for blur detection. CUHK \cite{shi2014discriminative} is a classical blur detection dataset, among which 296 images are partially motion-blurred, and 704 images are defocus-blurred. DUT \cite{zhao2018defocus} is a new defocus blur detection dataset which consists of 500 images as the test set and 600 images as the train set. We separate the CUHK blur dataset into a training set, which includes 800 images, and a test set, which includes 200 images that have the same ratio of motion-blurred images and defocus-blurred images. Since the number of training samples is limited, we enlarge training set by horizontal reversal at each orientation. Due to the fact that some state-of-the-art methods were designed solely for defocus blur detection, when we compare to these methods on the CUHK blur dataset we only use the 704 defocus-blurred images from CUHK and separate them into a training set which includes 604 images and a test set which includes 100 images. Our experiments were performed on these three datasets (CUHK, DUT, CUHK-defocus).
	
	\begin{figure*}[htbp]	
		\subfigure[]{
			\begin{minipage}{0.075\linewidth}	
				\centering
				\includegraphics[width=1.0\linewidth]{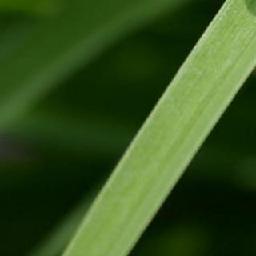}
				\includegraphics[width=1.0\linewidth]{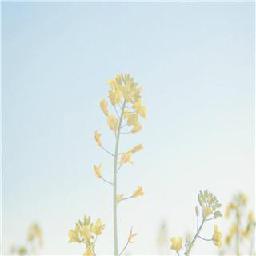}
				\includegraphics[width=1.0\linewidth]{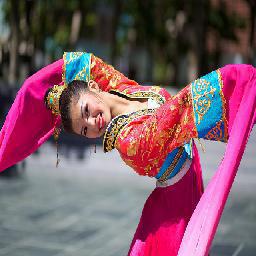}
				\includegraphics[width=1.0\linewidth]{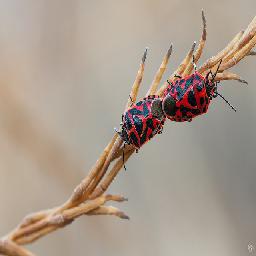}
				\includegraphics[width=1.0\linewidth]{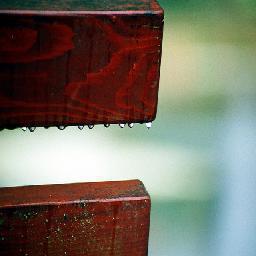}			
			\end{minipage}%
		}%
		\subfigure[]{
			\begin{minipage}{0.075\linewidth}	
				\centering
				\includegraphics[width=1.0\linewidth]{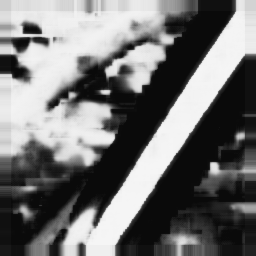}
				\includegraphics[width=1.0\linewidth]{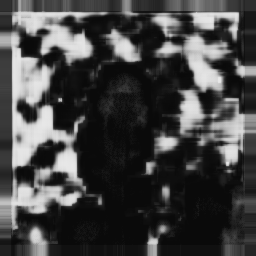}
				\includegraphics[width=1.0\linewidth]{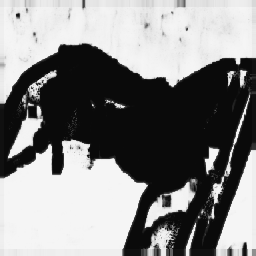}
				\includegraphics[width=1.0\linewidth]{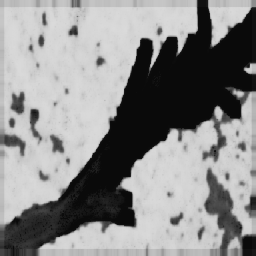}
				\includegraphics[width=1.0\linewidth]{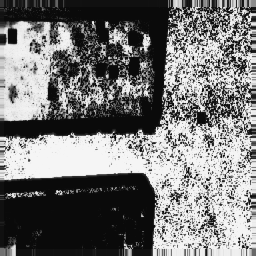}	
			\end{minipage}%
		}%
		\subfigure[]{
			\begin{minipage}{0.075\linewidth}	
				\centering
				\includegraphics[width=1.0\linewidth]{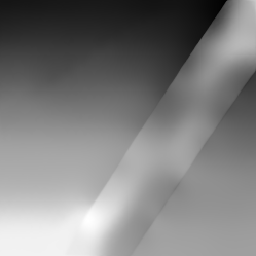}
				\includegraphics[width=1.0\linewidth]{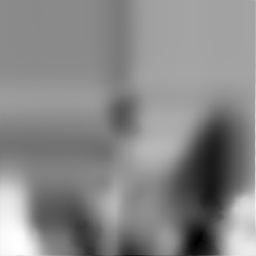}
				\includegraphics[width=1.0\linewidth]{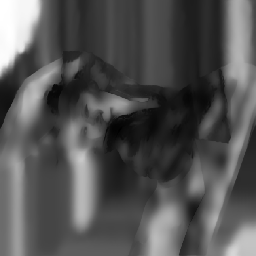}
				\includegraphics[width=1.0\linewidth]{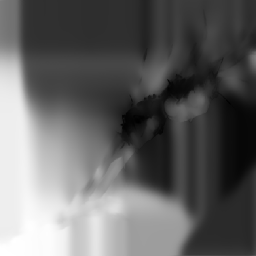}
				\includegraphics[width=1.0\linewidth]{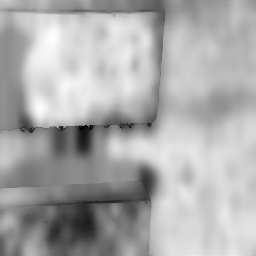}	
			\end{minipage}%
		}%
		\subfigure[]{
			\begin{minipage}{0.075\linewidth}	
				\centering
				\includegraphics[width=1.0\linewidth]{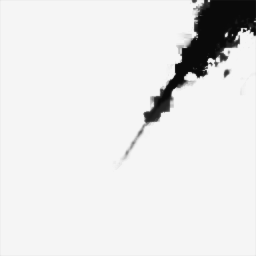}
				\includegraphics[width=1.0\linewidth]{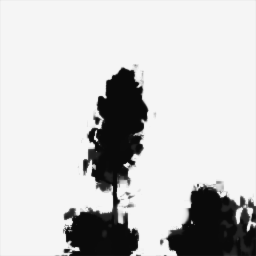}
				\includegraphics[width=1.0\linewidth]{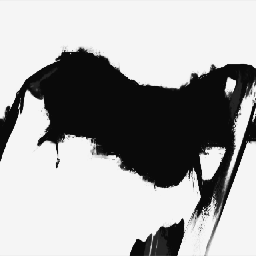}
				\includegraphics[width=1.0\linewidth]{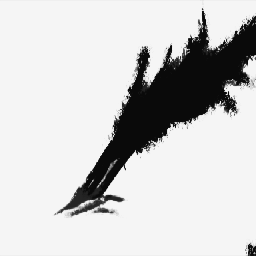}
				\includegraphics[width=1.0\linewidth]{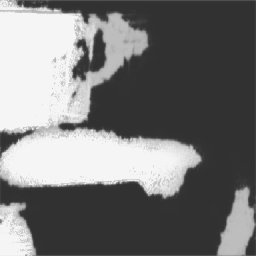}	
			\end{minipage}%
		}%
		\subfigure[]{
			\begin{minipage}{0.075\linewidth}	
				\centering
				\includegraphics[width=1.0\linewidth]{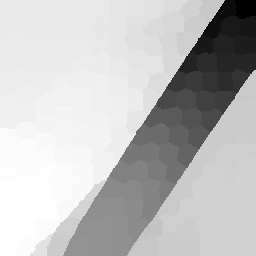}
				\includegraphics[width=1.0\linewidth]{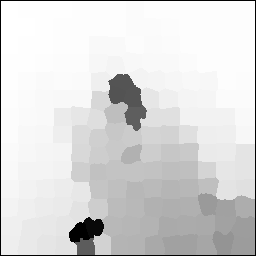}
				\includegraphics[width=1.0\linewidth]{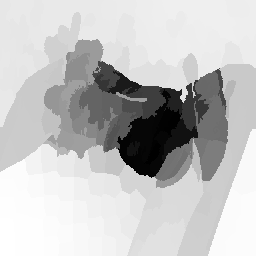}
				\includegraphics[width=1.0\linewidth]{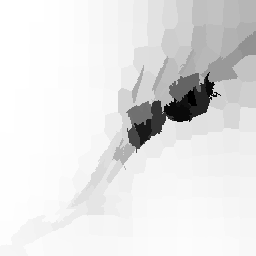}
				\includegraphics[width=1.0\linewidth]{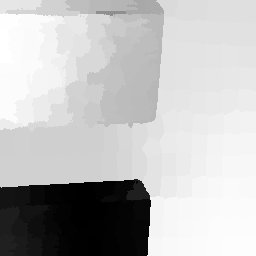}	
			\end{minipage}%
		}%
		\subfigure[]{
			\begin{minipage}{0.075\linewidth}	
				\centering
				\includegraphics[width=1.0\linewidth]{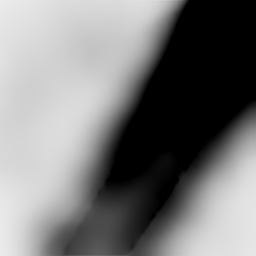}
				\includegraphics[width=1.0\linewidth]{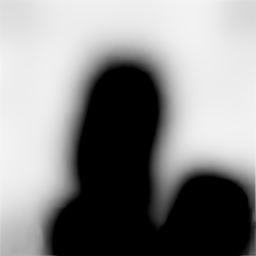}
				\includegraphics[width=1.0\linewidth]{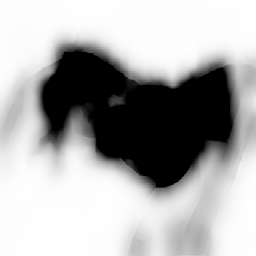}
				\includegraphics[width=1.0\linewidth]{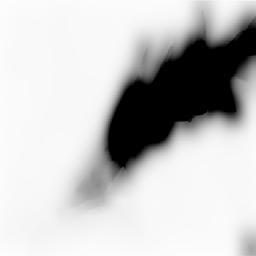}
				\includegraphics[width=1.0\linewidth]{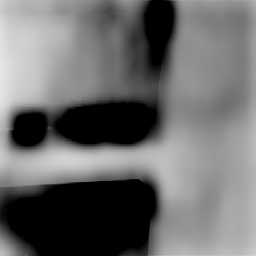}	
			\end{minipage}%
		}%
		\subfigure[]{
			\begin{minipage}{0.075\linewidth}	
				\centering
				\includegraphics[width=1.0\linewidth]{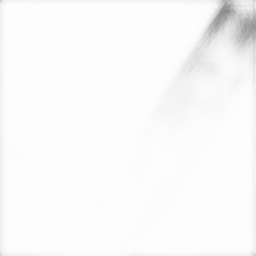}
				\includegraphics[width=1.0\linewidth]{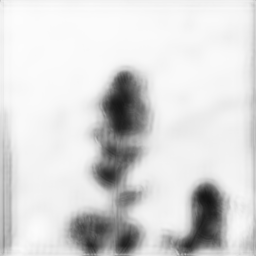}
				\includegraphics[width=1.0\linewidth]{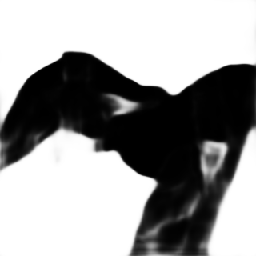}
				\includegraphics[width=1.0\linewidth]{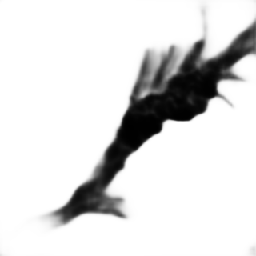}
				\includegraphics[width=1.0\linewidth]{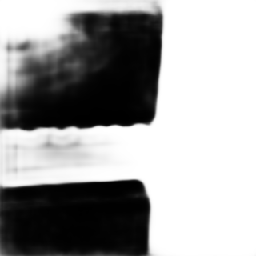}	
			\end{minipage}%
		}%
		\subfigure[]{
			\begin{minipage}{0.075\linewidth}	
				\centering
				\includegraphics[width=1.0\linewidth]{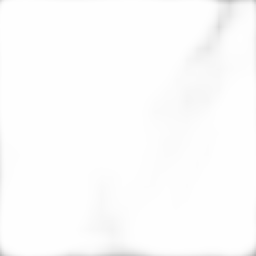}
				\includegraphics[width=1.0\linewidth]{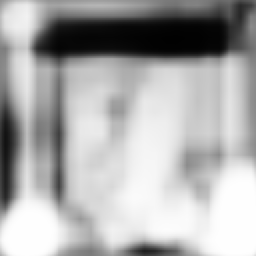}
				\includegraphics[width=1.0\linewidth]{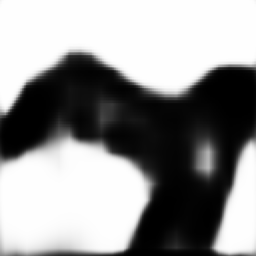}
				\includegraphics[width=1.0\linewidth]{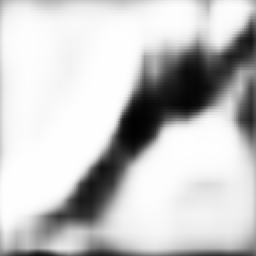}
				\includegraphics[width=1.0\linewidth]{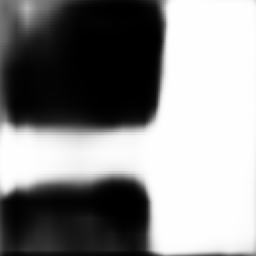}	
			\end{minipage}%
		}%
		\subfigure[]{
			\begin{minipage}{0.075\linewidth}	
				\centering
				\includegraphics[width=1.0\linewidth]{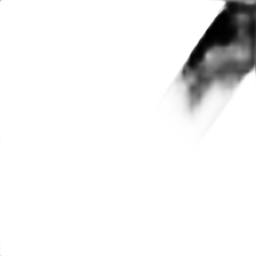}
				\includegraphics[width=1.0\linewidth]{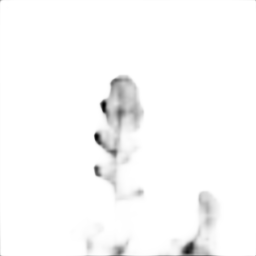}
				\includegraphics[width=1.0\linewidth]{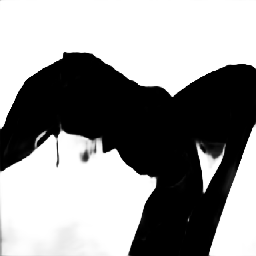}
				\includegraphics[width=1.0\linewidth]{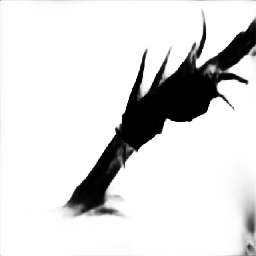}
				\includegraphics[width=1.0\linewidth]{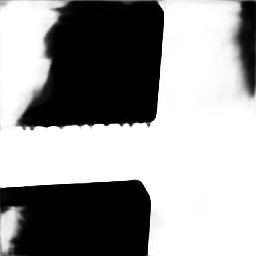}	
			\end{minipage}%
		}%
		\subfigure[]{
			\begin{minipage}{0.075\linewidth}	
				\centering
				\includegraphics[width=1.0\linewidth]{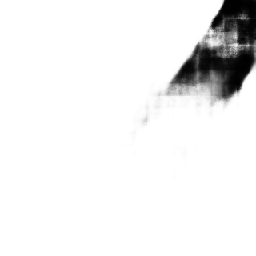}
				\includegraphics[width=1.0\linewidth]{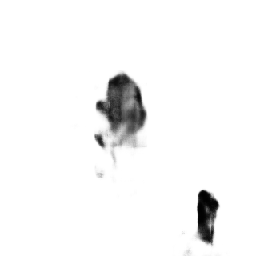}
				\includegraphics[width=1.0\linewidth]{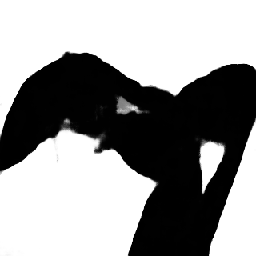}
				\includegraphics[width=1.0\linewidth]{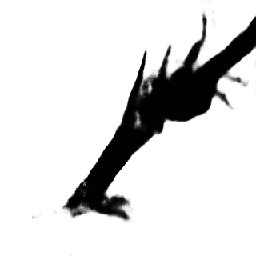}
				\includegraphics[width=1.0\linewidth]{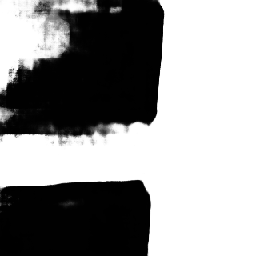}	
			\end{minipage}%
		}%
		\subfigure[]{
			\begin{minipage}{0.075\linewidth}	
				\centering
				\includegraphics[width=1.0\linewidth]{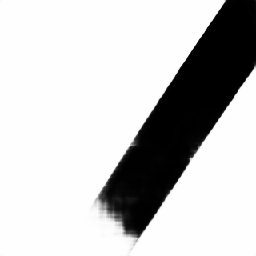}
				\includegraphics[width=1.0\linewidth]{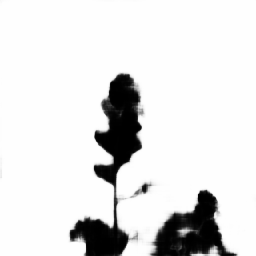}
				\includegraphics[width=1.0\linewidth]{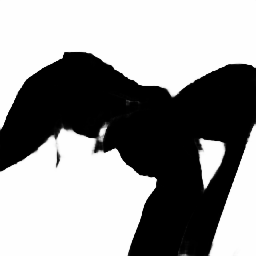}
				\includegraphics[width=1.0\linewidth]{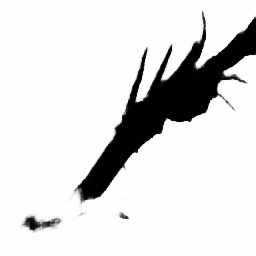}
				\includegraphics[width=1.0\linewidth]{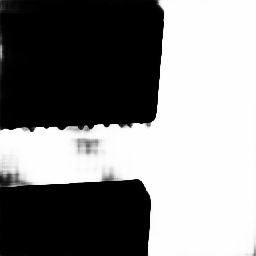}	
			\end{minipage}%
		}%
		\subfigure[]{
			\begin{minipage}{0.075\linewidth}	
				\centering
				\includegraphics[width=1.0\linewidth]{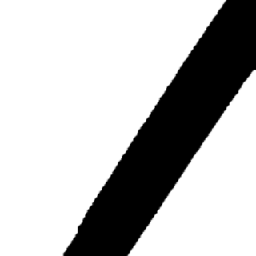}
				\includegraphics[width=1.0\linewidth]{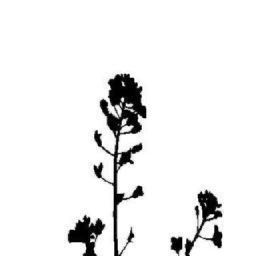}
				\includegraphics[width=1.0\linewidth]{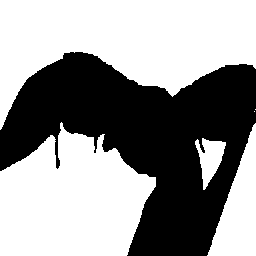}
				\includegraphics[width=1.0\linewidth]{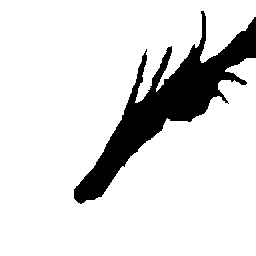}
				\includegraphics[width=1.0\linewidth]{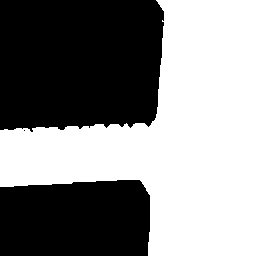}	
			\end{minipage}%
		}%
		
		\caption{ The defocus blur maps of generated from different methods. In the visual comparison, we can find that our method performer better in the scenes with similar background or cluttered background. (a) Input, (b) DBDF14, (c) JNB15, (d) LBP16, (e) SS16, (f) HiFST17, (g) BTB18, (h) DBM18, (i) DeF19, (j) CENet19, (k) Ours, (l) GT.}
		\label{fig5}
	\end{figure*}
	
	\begin{figure*}[htbp]
		\centering
		\subfigure[]{
			\begin{minipage}{0.12\linewidth}
				\centering
				\includegraphics[width= 1.0\linewidth]{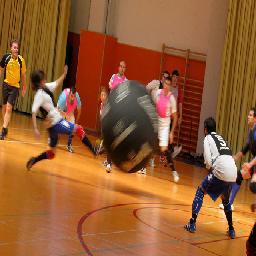}
				\includegraphics[width= 1.0\linewidth]{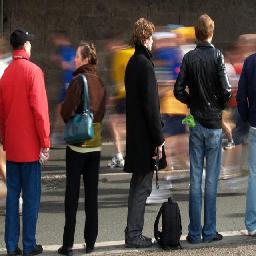}
			\end{minipage}
		}
		\subfigure[]{
			\begin{minipage}{0.12\linewidth}
				\centering
				\includegraphics[width= 1.0\linewidth]{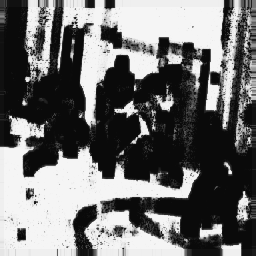}
				\includegraphics[width= 1.0\linewidth]{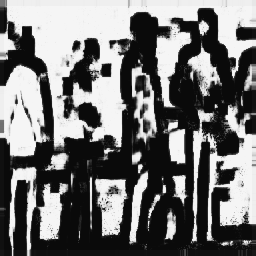}
			\end{minipage}
		}
		\subfigure[]{
			\begin{minipage}{0.12\linewidth}
				\centering
				\includegraphics[width= 1.0\linewidth]{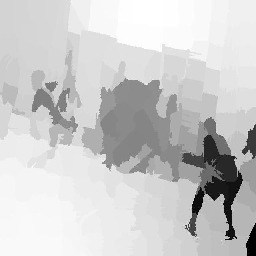}
				\includegraphics[width= 1.0\linewidth]{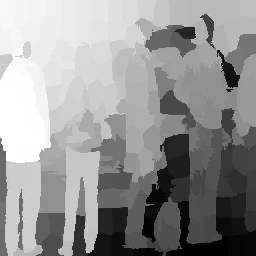}
			\end{minipage}
		}
		\subfigure[]{
			\begin{minipage}{0.11\linewidth}
				\centering
				\includegraphics[width= 1.0\linewidth]{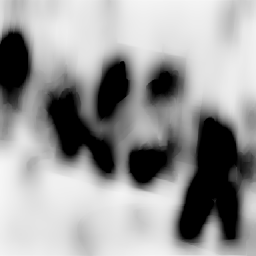}
				\includegraphics[width= 1.0\linewidth]{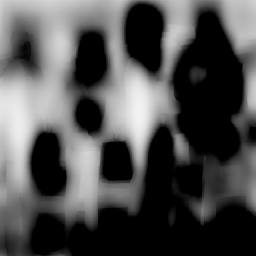}
			\end{minipage}
		}
		\subfigure[]{
			\begin{minipage}{0.11\linewidth}
				\centering
				\includegraphics[width= 1.0\linewidth]{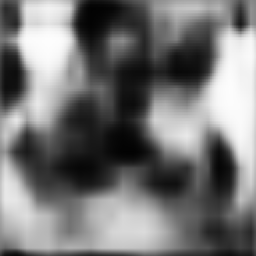}
				\includegraphics[width= 1.0\linewidth]{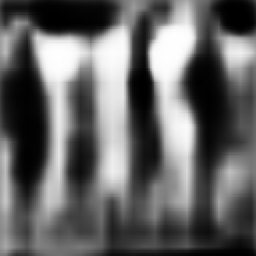}
			\end{minipage}
		}
		\subfigure[]{
			\begin{minipage}{0.11\linewidth}
				\centering
				\includegraphics[width= 1.0\linewidth]{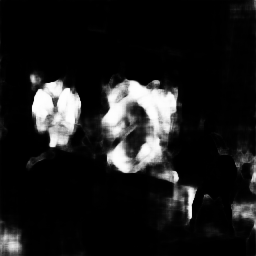}
				\includegraphics[width= 1.0\linewidth]{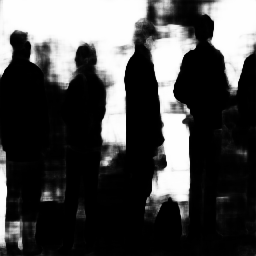}
			\end{minipage}
		}
		\subfigure[GT]{
			\begin{minipage}{0.11\linewidth}
				\centering
				\includegraphics[width= 1.0\linewidth]{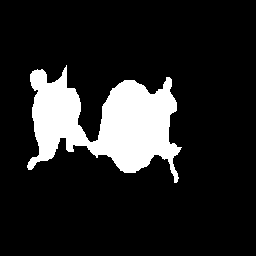}
				\includegraphics[width= 1.0\linewidth]{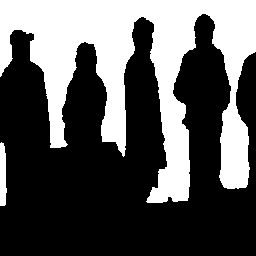}
			\end{minipage}
		}
		\caption{The motion blur maps of generated from different methods. In the visual comparison,our method make a giant progress than the other methods. (a) Input, (b) DBDF14, (c) SS16, (d) HiFST17, (e) DBM18, (f ) Ours, (g) GT.}
		\label{fig6}
	\end{figure*}
	\begin{figure*}[htbp]	
		\subfigure[DUT test set]{
			\begin{minipage}{0.33\linewidth}	
				\centering
				\includegraphics[width=1.0\linewidth]{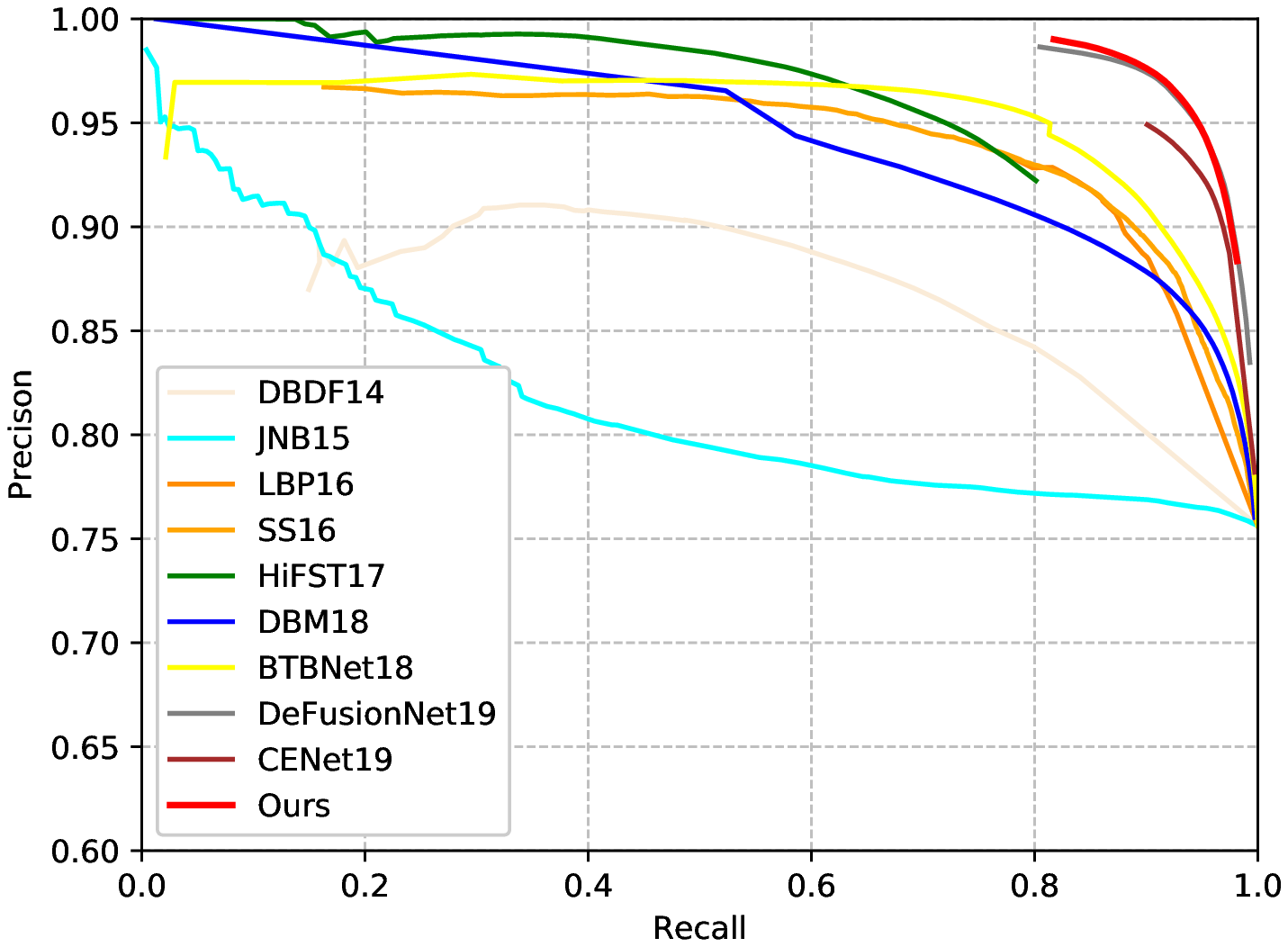}
			\end{minipage}%
		}%
		\subfigure[CUHK$^{*}$ test set(without motion blurred images)]{
			\begin{minipage}{0.33\linewidth}	
				\centering
				\includegraphics[width=1.0\linewidth]{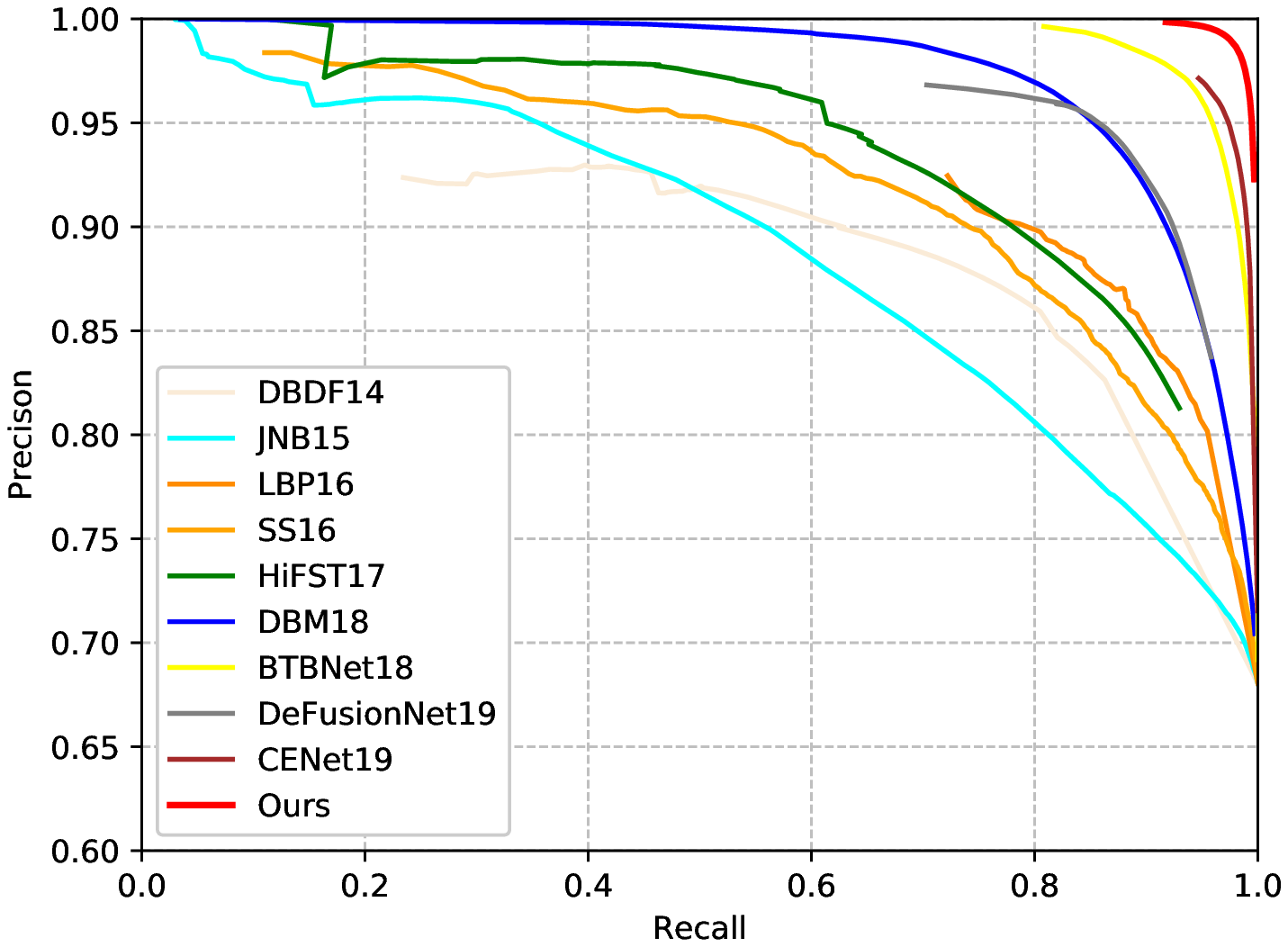}
			\end{minipage}%
		}%
		\subfigure[CUHK test set]{
			\begin{minipage}{0.33\linewidth}	
				\centering
				\includegraphics[width=1.0\linewidth]{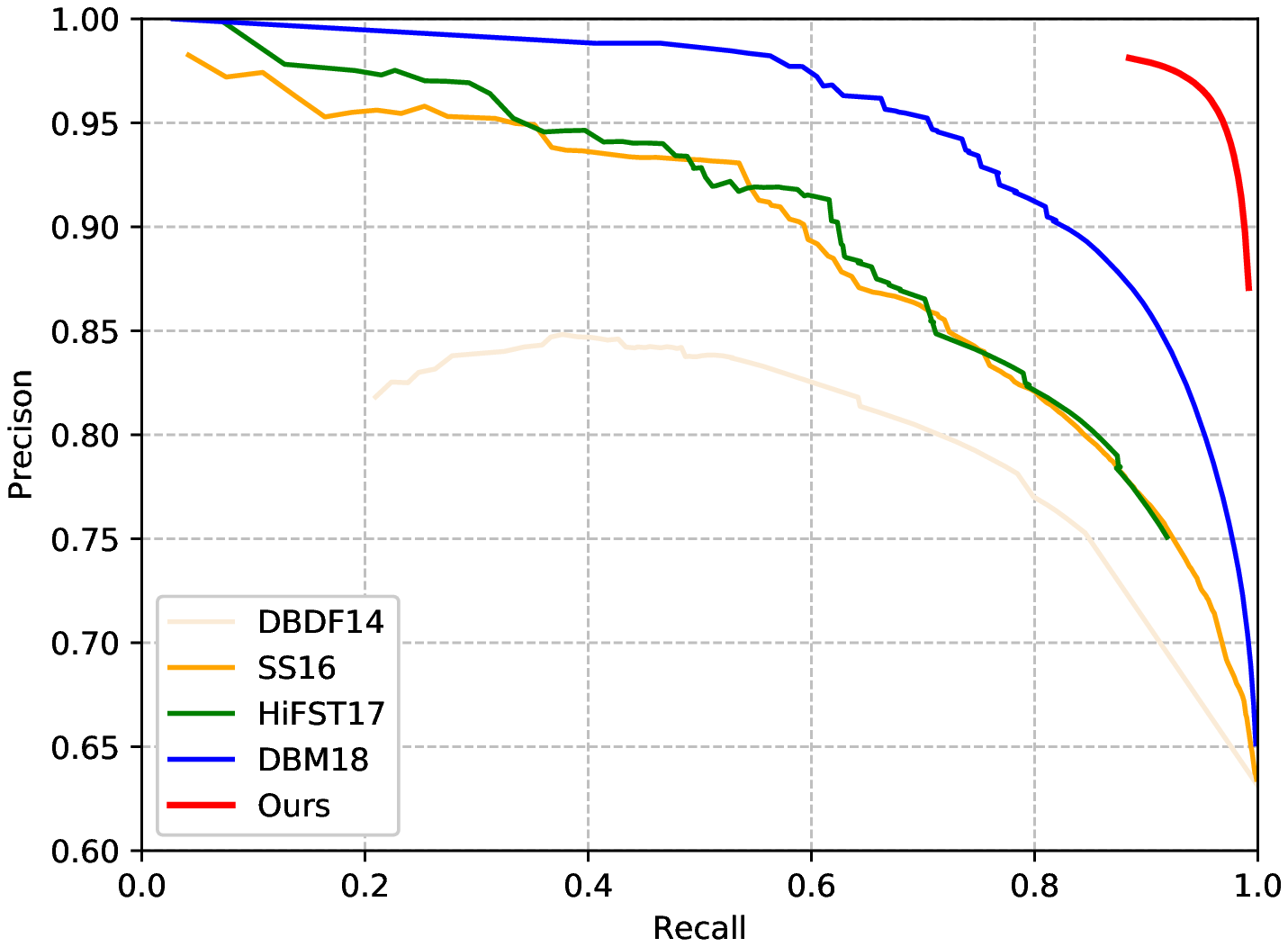}
			\end{minipage}%
		}%   	
		\caption{Comparison of precision-recall curves of different methods on three test sets. The curves of  our method are over $95\%$. Specially, our method have got more than 0.2 progress in precision than other method on the CUHK.}
		\label{fig7}
	\end{figure*}

	\begin{figure*}[htbp]	
	\subfigure[DUT test set]{
		\begin{minipage}{0.33\linewidth}	
			\centering
			\includegraphics[width=1.0\linewidth]{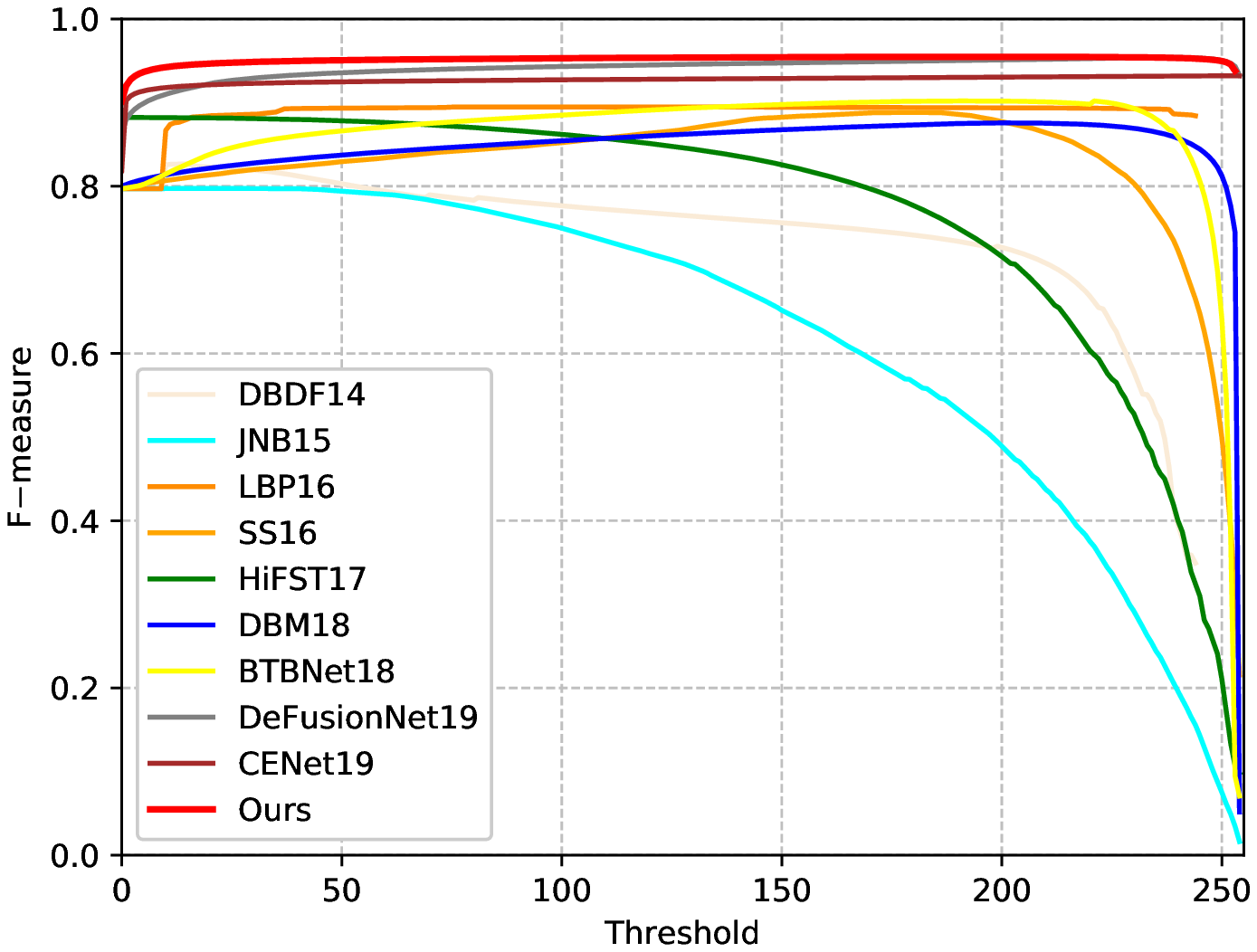}
		\end{minipage}%
	}%
	\subfigure[CUHK$^{*}$ test set(without motion blurred images)]{
		\begin{minipage}{0.33\linewidth}	
			\centering
			\includegraphics[width=1.0\linewidth]{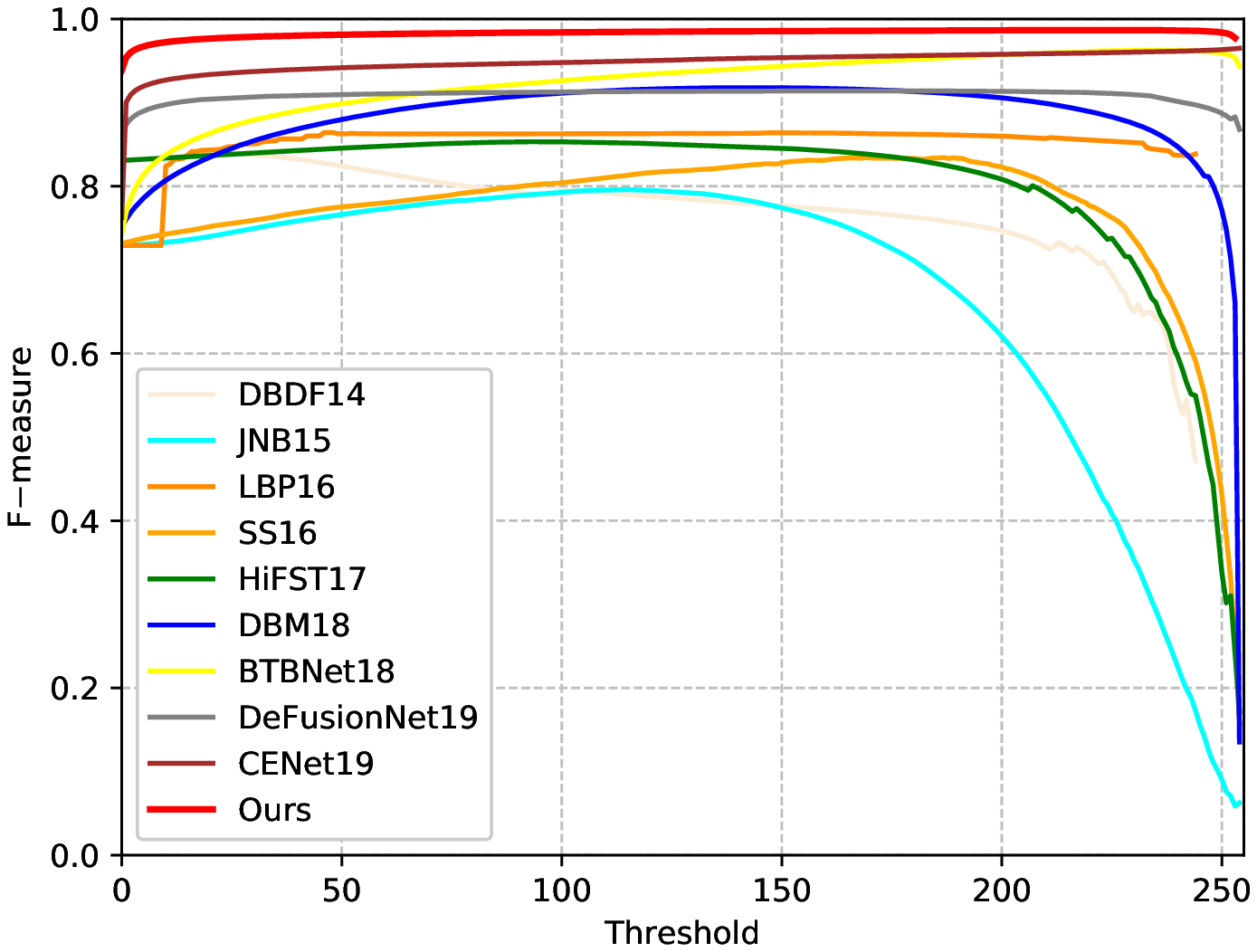}
		\end{minipage}%
	}%
	\subfigure[CUHK test set]{
		\begin{minipage}{0.33\linewidth}	
			\centering
			\includegraphics[width=1.0\linewidth]{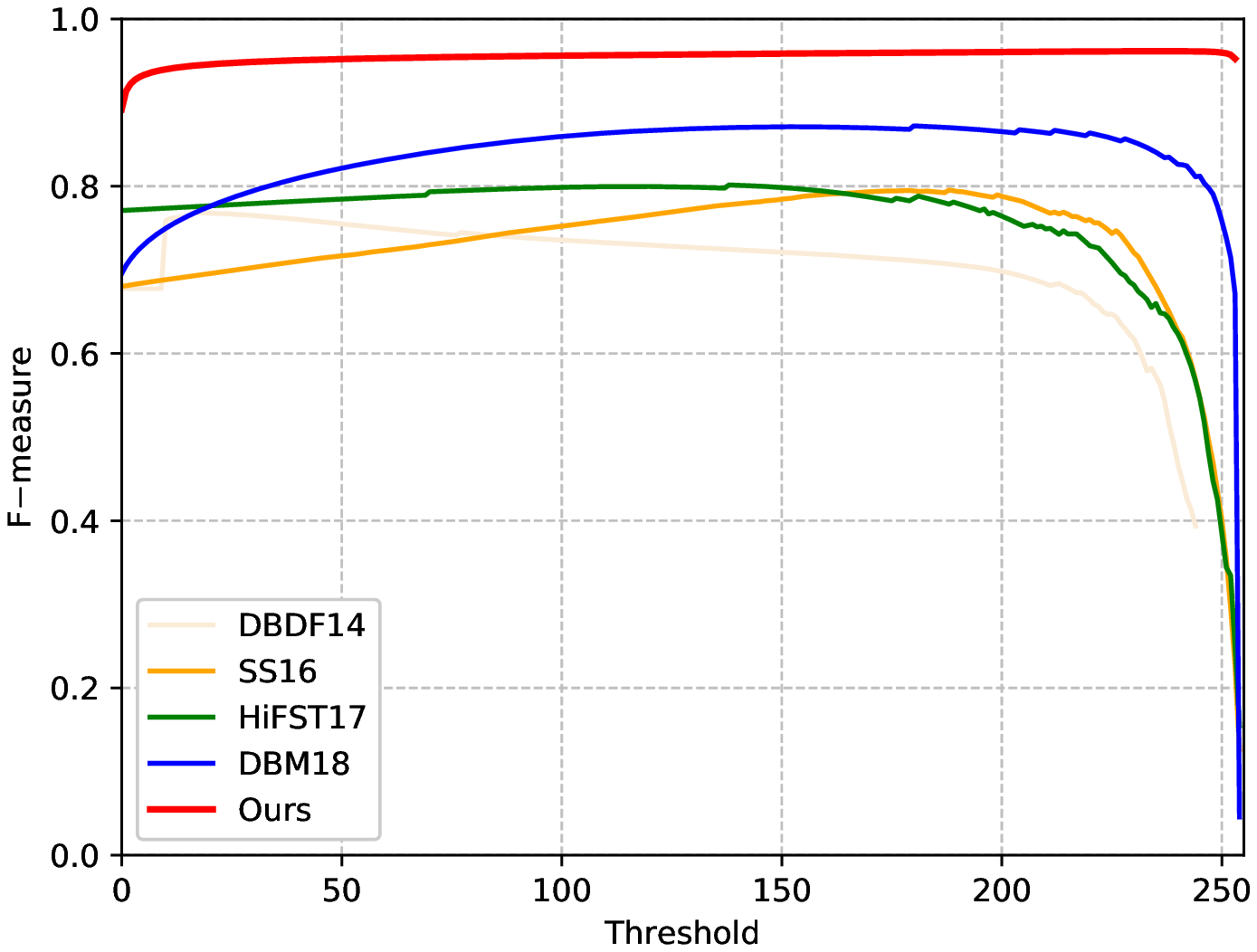}
		\end{minipage}%
	}%
	
	\caption{Comparison of F$_{\sqrt{0.3}}$-measure curves of different methods on three test sets. The curves of our method are the highest curves on the three test sets.}
	\label{fig8}
\end{figure*}
	
	We implement our model in Pytorch and train our model on a machine equipped with an Nvidia Tesla M40 GPU, in which the memory is 12 GB. We optimise the network by using Stochastic Gradient Gescent (SGD) algorithm with the momentum of $0.9$, the weight decay of $5e^{-4}$ and the learning rate of $0.01$ in the beginning and reduced by a factor of 0.1 every 25 epochs. We trained with a batch size of 16 and resize the input images' size as $256\time256$, which required $10 GB$ of GPU memory for training. We use our enhanced training set of 5200 images to train our model for a total of 100 epochs, which takes five hours.

	\subsection{Evaluation Criteria and Comparison}
	
	\textbf{Precision and Recall.}
	We vary the threshold used to produce a segmentation of sharpness maps to draw a curve.
	\begin{equation}
	precision=\frac{\mathbf{R} \cap \mathbf{R}_{g}}{\mathbf{R}}\ \ , \ recall\ =\frac{\mathbf{R} \cap \mathbf{R}_{g}}{\mathbf{R}_{g}}
	\end{equation}
	where $\mathbf{R}$ is the set of pixels in the segmented blurred region and $\mathbf{R}_{g}$ is the set of pixels in the ground truth blurred. The threshold $T_{seg}$ value is sampled at every integer within the interval $[0, 255]$.
	
	\textbf{F-measure.}
	The F-measure, which is an overall performance measurement, is defined as:
	\begin{equation}
	F_{\beta } =\frac{\left( 1+\beta ^{2}\right) \cdotp precision\cdotp recall}{\beta ^{2} \cdotp precision+recall}
	\end{equation}
	The $\beta$ is the weighting parameter ranging from 0 to 1. There $\beta^{2}=0.3$ following \cite{yi2016lbp} is employed to emphasize the precision. Precision stands for the percentage of sharp pixels being correctly detected, and Recall is the fraction of detected sharp pixels in relation to the ground truth number of sharp pixels. A larger $F$ value means a better result.
	
	\textbf{Mean Absolute Error(MAE).}
	MAE can provide a good measure of the dissimilarity between ground truth and blurred map.
	\begin{equation}
	\mathit{MAE}=\frac{1}{W\cdotp H}\sum^{W}_{x=1}\sum ^{H}_{y\ =1}|\mathbf{G}( x, y) -\mathbf{M}_{final}(x, y)|
	%MAE=\frac{1}{W\times H}\sum^{W}_{x=1}\sum ^{H}_{y\ =1}|\mathbf{G}( x, y) -\mathbf{M}_{final}(x, y)|
	\end{equation}
	where $x, y$ stand for pixel coordinates. $\mathbf{G}$ is the ground truth map, and $\mathbf{M}_{final}$ is the result map. $W $ and $H$ denote the width and height of the $\mathbf{M}_{final}$ (or $\mathbf{G}$), respectively. A smaller MAE value usually means a more accurate result.
	
	\begin{table*}
		\begin{center}
			\begin{tabular}[l]{@{}lccccccccccc}
				\hline
				Datasets & Metric & DBDF & JNB & LBP & SS & HiFST & DBM & BTB & DeF & CENet & Ours \\
				\hline\hline
				\multirow{2}*{DUT}
				& F$_{\sqrt{0.3}}$ & 0.827 & 0.798 & 0.895 & 0.889 & 0.883 & 0.876 & 0.902 & 0.953 & 0.932 & \textbf{0.954}\\
				~ & MEA & 0.244 & 0.244 & 0.168 & 0.163 & 0.203 & 0.165 & 0.145 & 0.078 & 0.098 &\textbf{0.075} \\
				\hline
				\multirow{2}*{CUHK$^{*}$}
				& F$_{\sqrt{0.3}}$ &0.841&0.796&0.864&0.834&0.853&0.918&0.963&0.914& 0.965 &\textbf{0.976} \\
				~ &MEA&0.208&0.260&0.174&0.215&0.179&0.114&0.057&0.103& 0.049 &\textbf{0.032}  \\
				\hline
				\multirow{2}*{CUHK}
				& F$_{\sqrt{0.3}}$ &0.768&-&-&0.795&0.799& 0.871 &-&-&-&\textbf{0.953} \\
				~&MEA&0.257&-&-&0.248&0.207&0.123&-&-&-& \textbf{0.042}\\
				\hline
			\end{tabular}
		\end{center}
		\caption{Quantitative comparison of F$_{\sqrt{0.3}}$-measure and MEA scores. The best results are marked out in bold. CUHK$^{*}$ in the table is the CUHK dataset excluding motion-blurred images, and "-" means that the methods are not designed for the motion blur. }
		\label{tab1}
	\end{table*}
	
	We compare our method against other 9 state-of-the-art methods, including deep learning-based methods and hand-crafted features methods: DeF \cite{tang2019defusionnet}, CENet \cite{zhao2019enhancing}, BTBNet \cite{zhao2018defocus}, DBM \cite{ma2018deep}, HIFST \cite{golestaneh2017spatially}, SS \cite{tang2016spectral}, LBP \cite{yi2016lbp}, JNB \cite{shi2015just}, and DBDF \cite{shi2014discriminative}. In Figure \ref{fig5}, we show some defocus-blurred cases of visual comparison results. These cases include various scenes with cluttered backgrounds or similar backgrounds and contain complex boundaries of objects, which are hard to separate the sharp regions from images. In Figure \ref{fig6}, we show some motion-blurred cases of visual comparison result of different methods that can be applied in motion blur detection.

	 We also draw the accurate precision-recall curves and F-measure curves to study the capabilities of these methods through statistical calculation. In Figure \ref{fig7}, it is shown that our method makes progress on all three tests, and especially on the CUHK dataset which contains both defocus-blurred images and motion-blurred images. Our method boosts the precision within the entire recall range, where the improvement can be as large as $0.2$. Furthermore, in Figure \ref{fig8}, the F-measure curves of our methods are all over $0.9$, which are the best on each dataset.

	In Table \ref{tab1},  it is observed that our method consistently performs favourably against other methods on three data sets, which indicates the superiority of our method over other approaches.	
	
	\subsection{Ablation Analysis}
	\textbf{Effectiveness of Skip layers.}
	
    \begin{table}
		\begin{center}
			\begin{tabular}[l]{@{}lcc}
				\hline
				Network & No Skip & Ours \\
				\hline\hline
				F$_{\sqrt{0.3}}$-measure & 0.851 & 0.952 \\
				\hline
				MEA & 0.137 & 0.042 \\
				\hline
			\end{tabular}
		\end{center}
	    \caption{Quantitative comparison of F$_{\sqrt{0.3}}$-measure and MEA scores between our model and the model without skip connections.}
		\label{tab2}
	\end{table}
	\begin{figure}[htbp]	
		\subfigure[Source]{
			\begin{minipage}{0.25\linewidth}	
				\centering
				\includegraphics[width=1.0\linewidth]{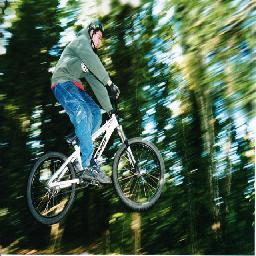}
				\includegraphics[width=1.0\linewidth]{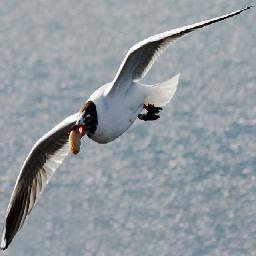}
			\end{minipage}%
		}%
		\subfigure[No-Skips]{
			\begin{minipage}{0.25\linewidth}	
				\centering
				\includegraphics[width=1.0\linewidth]{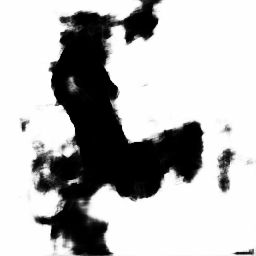}
				\includegraphics[width=1.0\linewidth]{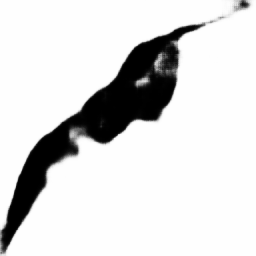}
			\end{minipage}%
		}%
		\subfigure[Ours]{
			\begin{minipage}{0.25\linewidth}	
				\centering
				\includegraphics[width=1.0\linewidth]{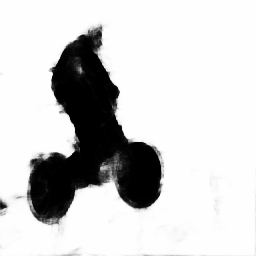}
				\includegraphics[width=1.0\linewidth]{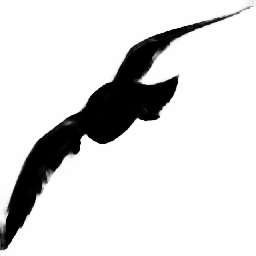}
			\end{minipage}%
		}%
		\subfigure[GT]{
			\begin{minipage}{0.25\linewidth}	
				\centering
				\includegraphics[width=1.0\linewidth]{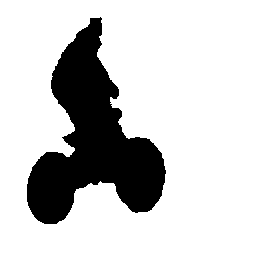}
				\includegraphics[width=1.0\linewidth]{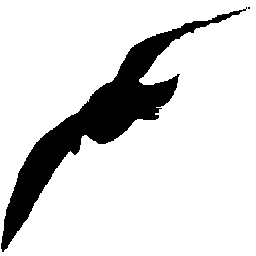}
			\end{minipage}%
		}%
		\caption{Visual comparison results between ours and the model without skip connections.}
		\label{fig9}
	\end{figure}

	Although U-shaped networks with skip layers have been applied in BTBNet, we make supplementary experiments to verify the significance of skip connections. To control variable, we build a new model that is similar with our original model except that there are no skip layers, using CUHK blur dataset for training. By the comparison of the result, we find that the model without skip connections cannot precisely segment the edges of objects in Figure \ref{fig9}. As a result, the model skip connections has a lower F$_{\sqrt{0.3}}$-measure score and a higher MEA score in Table \ref{tab2}.

	\textbf{Effectiveness of multi-scale Extractors.}
	\begin{table}
		\begin{center}
			\begin{tabular}[l]{@{}lccc}
				\hline
				Network & U-net & No dilated($5\times5$) &  Ours  \\
				\hline\hline
				F$_{\sqrt{0.3}}$-measure & 0.843 & 0.956  & 0.950 \\
				\hline
				MEA & 0.146 & 0.044 & 0.046 \\
				\hline
			\end{tabular}
		\end{center}
		\caption{Quantitative comparison of F$_{\sqrt{0.3}}$-measure and MEA scores among no dilatited (using $5\times5$ normal convolution kernels), ours and U-net.}
		\label{tab3}
	\end{table}
\begin{figure}[htbp]	
	\subfigure[Source]{
		\begin{minipage}{0.20\linewidth}	
			\centering
			\includegraphics[width=1.0\linewidth]{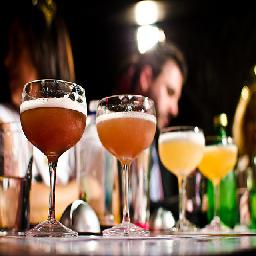}
			\includegraphics[width=1.0\linewidth]{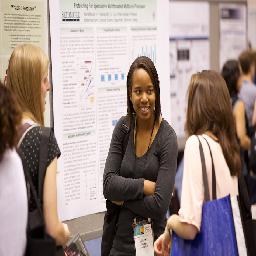}
		\end{minipage}%
	}%
	\subfigure[U-net]{
		\begin{minipage}{0.20\linewidth}	
			\centering
			\includegraphics[width=1.0\linewidth]{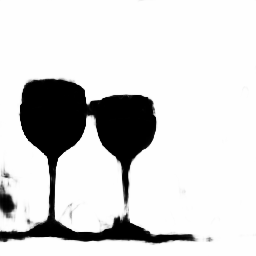}
			\includegraphics[width=1.0\linewidth]{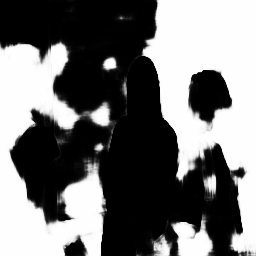}
		\end{minipage}%
	}%
	\subfigure[No dilated($5\times5$)]{
		\begin{minipage}{0.20\linewidth}	
			\centering
			\includegraphics[width=1.0\linewidth]{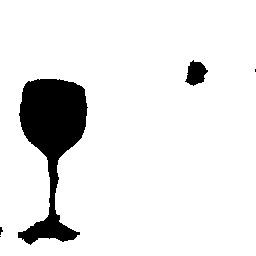}
			\includegraphics[width=1.0\linewidth]{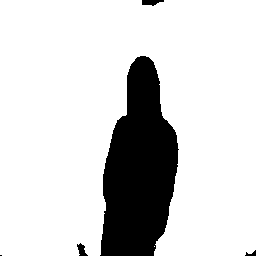}
		\end{minipage}%
	}%
	\subfigure[Ours]{
		\begin{minipage}{0.20\linewidth}	
			\centering
			\includegraphics[width=1.0\linewidth]{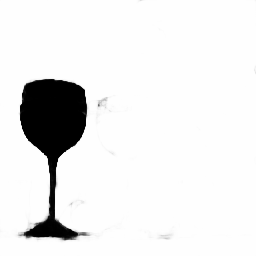}
			\includegraphics[width=1.0\linewidth]{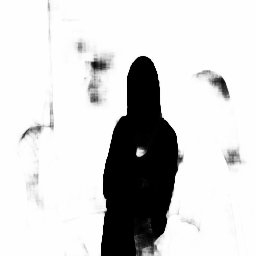}
		\end{minipage}%
	}%
	\subfigure[GT]{
		\begin{minipage}{0.20\linewidth}	
			\centering
			\includegraphics[width=1.0\linewidth]{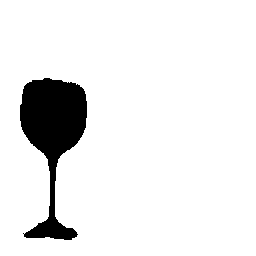}
			\includegraphics[width=1.0\linewidth]{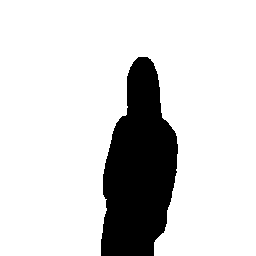}
		\end{minipage}%
	}%
	\caption{Visual comparison results among no dilatited (using $5\times5$ normal convolution kernels), ours and U-net.}
	\label{fig10}
\end{figure}
    Multi-scale Extractors with dilated convolution aim to extract multi-scale texture feature to improve the precision of the blurred map. To verify its effect, we compare our network with the classical U-net which does not have multi-scale extractors and resembles our network. In Figure \ref{fig10}, we can find that the results of the U-net \cite{ronneberger2015u} without the multi-scale extractors are disturbed by backgrounds of shallow depths. Owing to the multi-scale extractors, our model is so sensitive to degree of blur that can accurately separate the blur region. As a result, our model has a higher F$_{\sqrt{0.3}}$-measure score and a lower MEA score in Table \ref{tab3}. Also, we have a try to replace $3\times3$ 1-dilated convolution kernels with the $5 \times5$ normal convolution kernels, which have the same receptive field. However, as shown in Table \ref{tab3}, our model performs a bit worse than the model using $5\times5$ normal convolution kernels. But our model save millions of parameters by using dilated convolutions.

	\section{Conclusions}
	
	In this work, we regard the blur detection as image segmentation. We design a group of multi-scale extractors with dilated convolution to capture different scale texture information of images. Then, we combined the extractors with the U-shape network to fuse the shallow texture information and the deep semantic information. Taking the advantage of the multi-scale texture information and the semantic information, our method performer better on the scenes with cluttered backgrounds or similar backgrounds and contain complex boundaries of objects. We test our model on three datasets. Experimental results on three datasets prove that our method performs better than the state-of-the-art methods in blur detection. Although our method have made progress, the performance of our method is limited by the richness of the dataset. In the future, we will make further study to improve the generalization of our model.

    \bibliographystyle{splncs04}
    \bibliography{egbib}

\end{document}